\documentclass[10pt,twocolumn,letterpaper]{article}

\usepackage{cvpr}
\usepackage{times}
\usepackage{epsfig}
\usepackage{graphicx}
\usepackage{amsmath}
\usepackage{amssymb}

\usepackage{epsfig}
\usepackage{amsmath}
\usepackage{amssymb}
\usepackage{color}
\usepackage{comment}
\usepackage{algorithm}
\usepackage{algorithmic}
\usepackage{multirow}
\usepackage{rotating}
\usepackage[pagebackref=true,breaklinks=true,letterpaper=true,colorlinks,bookmarks=false]{hyperref}

\cvprfinalcopy 

\def\cvprPaperID{2349} 

\ifcvprfinal\fi
\begin{document}

\title{Interpreting CNNs via Decision Trees}

\author{Quanshi Zhang$^{\dag}$, Yu Yang$^{\ddag}$, Haotian Ma$^{\S}$, and Ying Nian Wu$^{\ddag}$\\
$^{\dag}$Shanghai Jiao Tong University, \quad$^{\ddag}$University of California, Los Angeles,\\$^{\S}$South China University of Technology
}

\maketitle

\begin{abstract}
This paper\footnote[1]{Quanshi Zhang is the corresponding author with the John Hopcroft Center and the MoE Key Lab of Artificial Intelligence, AI Institute, at the Shanghai Jiao Tong University, China. Yu Yang and Ying Nian Wu are with the University of California, Los Angeles, USA. Haotian Ma is with the South China University of Technology, China.} aims to quantitatively explain rationales of each prediction that is made by a pre-trained convolutional neural network (CNN). We propose to learn a decision tree, which clarifies the specific reason for each prediction made by the CNN at the semantic level. \emph{I.e.} the decision tree decomposes feature representations in high conv-layers of the CNN into elementary concepts of object parts. In this way, the decision tree tells people which object parts activate which filters for the prediction and how much they contribute to the prediction score. Such semantic and quantitative explanations for CNN predictions have specific values beyond the traditional pixel-level analysis of CNNs. More specifically, our method mines all potential decision modes of the CNN, where each mode represents a common case of how the CNN uses object parts for prediction. The decision tree organizes all potential decision modes in a coarse-to-fine manner to explain CNN predictions at different fine-grained levels. Experiments have demonstrated the effectiveness of the proposed method.
\end{abstract}

\begin{figure}[t]
\centering
\includegraphics[width=0.95\linewidth]{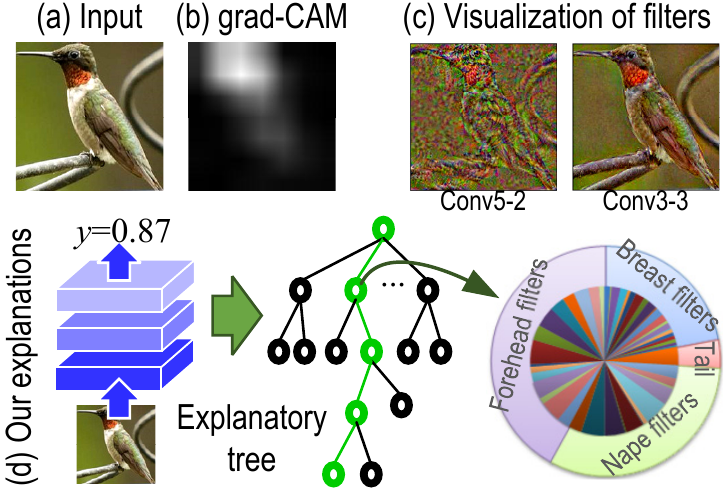}
\caption{Different types of explanations for CNNs. We compare (d) our task of quantitatively and semantically explaining CNN predictions with previous studies of interpreting CNNs, such as (b) the grad-CAM~\cite{visualCNN_grad_2} and (c) CNN visualization~\cite{CNNVisualization_2}. Given an input image (a), we infer a parse tree (green lines) within the decision tree to project neural activations onto clear concepts of object parts. Our method quantitatively explains which filters/parts (in the small/big round) are used for the prediction and how much they contribute to the prediction. We visualize numerical contributions from randomly selected 10\% filters for clarity.}
\label{fig:task}
\end{figure}

\section{Introduction}

Convolutional neural networks (CNNs)~\cite{CNN,CNNImageNet,ResNet} have achieved superior performance in various tasks. However, besides the discrimination power, model interpretability is still a significant challenge for neural networks. Many studies have been proposed to visualize or analyze feature representations hidden inside a CNN, in order to open the black box of neural networks.

\textbf{Motivation \& objective:} In the scope of network interpretability, state-of-the-art algorithms are still far from the ultimate goal of \textit{explaining why a CNN learns knowledge as it is}. Although some theories, such as the information bottleneck~\cite{InformationBottleneck}, analyzed statistical characteristics of a neural network, it is still a challenge to explain why a CNN encodes frontal-leg features, rather than rear-leg features, for classification during the end-to-end learning process.

Therefore, in this study, we limit our discussion to the issue of \textit{explaining what knowledge a CNN learns.} In this direction, our research focuses on the following two new perspectives of interpreting CNNs:
\begin{itemize}
\item How to explain features of a middle layer in a CNN at the semantic level. \emph{I.e.} we aim to transform chaotic features of filters inside a CNN into semantically meaningful concepts, such as object parts, so as to help people to understand the knowledge in the CNN.
\item How to quantitatively analyze the rationale of each CNN prediction. We need to figure out which filters/parts pass their appearance information through the CNN and contribute to the prediction output. We also report the numerical contribution of each filter (or object part) to the output score.
\end{itemize}
As shown in Fig.~\ref{fig:task}, above two perspectives are crucial in real applications and have essential differences from traditional pixel-level visualization and diagnosis of CNN features~\cite{CNNVisualization_1,CNNVisualization_2,FeaVisual,trust,shap}. Our semantic and quantitative explanations for CNNs have potential values beyond pixel-level visualization/analysis of CNNs. Semantic and quantitative explanations can help people better understand and trust CNN predictions. \emph{E.g.} in critical applications, such as the recommendation for a surgery plan, people are usually not simply satisfied by the plan itself, but expect a quantitative explanation for the plan.

\begin{figure}[t]
\centering
\includegraphics[width=0.95\linewidth]{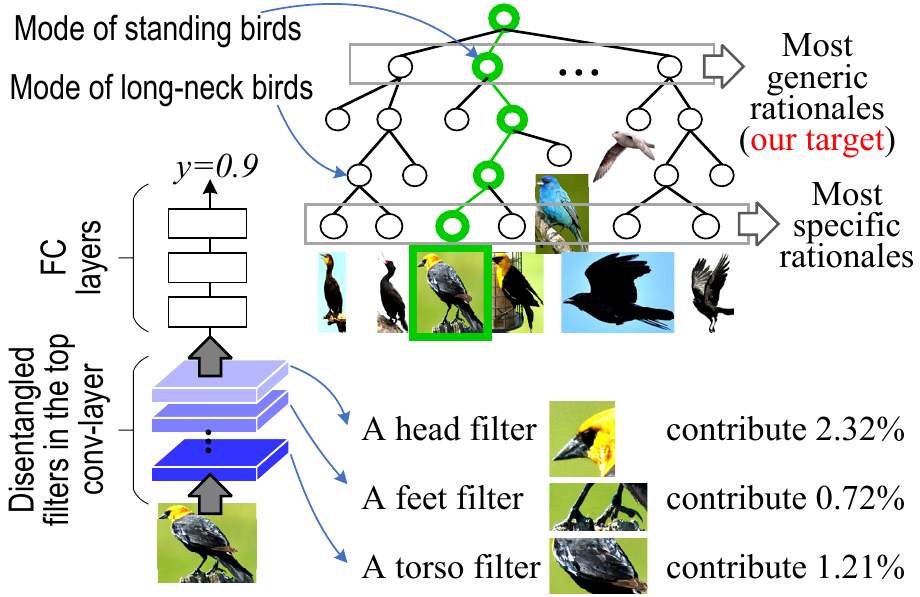}
\caption{Decision tree that encodes all potential decision modes of the CNN in a coarse-to-fine manner. We learn a CNN for object classification with disentangled representations in the top conv-layer, where each filter represents an object part. Given an input image, we infer a parse tree (green lines) from the decision tree to semantically and quantitatively explain which object parts (or filters) are used for the prediction and how much an object part (or filter) contributes to the prediction. We are more interested in high-layer decision modes that summarize low-layer modes into compact explanations of CNN predictions.}
\label{fig:decision}
\end{figure}

However, bridging the gap between a CNN's middle-layer features and semantic explanations has not been well explored yet. We will introduce our task, clarify its challenges, and define relevant concepts from the following two perspectives.
\begin{enumerate}
\item[1.] \textit{Bridging middle-layer features with semantic concepts:} Given an input image, the first issue is to learn more interpretable feature representations inside a CNN and associate each neural activation inside a CNN with a semantic concept. This presents significant challenges for state-of-the-art algorithms.
\end{enumerate}
Firstly, we need to force feature representations in middle conv-layers to be well disentangled during the learning process. According to \cite{Interpretability,explanatoryGraph}\footnote[2]{Zhang \emph{et al.}~\cite{interpretableCNN} summarized the six types of semantics defined in \cite{Interpretability} as parts and textures.}, a filter in traditional CNNs usually represents a mixture of parts and textures. Learning semantically meaningful filters is difficult but is the foundation of semantic-level explanations. In this research, we learn a CNN with disentangled filters in high conv-layers. Each filter needs to be consistently activated by the same object region over different input images. We do not use any annotations of parts or textures to supervise the disentanglement of filter features.

Secondly, we also need to associate each disentangled filter with an explicit semantic meaning (\emph{i.e.} an object part in this study). This enables linguistic descriptions of middle-layer knowledge, for example, how many parts are memorized in the CNN and how the parts are organized.
\begin{enumerate}
\item[2.] \textit{Bridging middle-layer features with final CNN predictions:} When we have assigned middle-layer features with specific part concepts, the next issue is to quantitatively explain how the CNN uses these middle-layer features to compute prediction scores. In other words, given an input image, we hope to clarify the specific rationale of the CNN prediction.
\end{enumerate}
Here, we define the {\bf\textit{rationale}} of a CNN prediction as the set of object parts (or filters) that are activated and contribute to the prediction. Given different input images, the CNN uses different object parts to activate different sets of filters to compute prediction scores, thereby having different rationales. Let us take the bird classification as an example. The CNN may use several filters activated by the head appearances as rationales to classify a standing bird, and the CNN may take filters for wings to distinguish a flying bird.

Given each input image, our task is to clarify filters of which object parts are activated and to quantitatively measure the contribution of each object part to the prediction. The concept of the contribution was also called the ``importance'' in \cite{trust,shap} and was termed ``attribution'' in \cite{patternNet}. As shown in Fig.~\ref{fig:decision}, we describe the contribution as ``a head filter contributes 2.32\%, and a feet filter contributes 0.72\%.''

\textbf{Task:} As shown in Fig.~\ref{fig:task}, given a pre-trained CNN, we propose a method to construct a decision tree to explain CNN predictions semantically and quantitatively. We summarize rationales of CNN predictions on all images into various {\bf\textit{decision modes}}. Each tree node represents a decision mode. Each decision mode describes common rationales of predictions that are shared by multiple images. \emph{I.e.} for these images, the CNN usually activates similar filters (object parts), and each part makes a similar contribution to the prediction.

The decision tree hierarchically represents all potential decision modes of a CNN in a coarse-to-fine manner. Nodes near to the tree root node mainly represent most common decision modes (prediction rationales) shared by many images. Nodes near leaves correspond to fine-grained modes of minority images. In particular, each leaf node encodes the specific decision mode of a certain image.

In order to build the decision tree, we learn filters to represent object parts (we do not label any parts or textures as additional supervision\footnote[3]{Part annotations are not used to learn the CNN and the decision tree. Given the learned CNN, we label object parts for the filters to compute part-level contributions in Equation (\ref{eqn:contri}).}). Then, we assign each filter with a certain part name. Finally, we mine decision modes to explain how the CNN use parts/filters for prediction, and construct a decision tree.

\textbf{Inference:} When the CNN makes a prediction for an input image, the decision tree determines a parse tree (see green lines in Fig.~\ref{fig:decision}) to encode a series of explanations. Each node (decision mode) in the parse tree quantitatively explains the prediction at a certain abstraction level, \emph{i.e.} clarifying how much each object part/filter contributes to the prediction score.

\textit{Compared to fine-grained modes in leave nodes, we are more interested in generic decision modes in high-level nodes. Generic decision modes usually select significant object parts (filters) as the rationale of CNN predictions and ignore insignificant ones. Thus, generic decision modes reflect compact rationales for CNN predictions.}

\textbf{Contributions:} In this paper, we aim to bridge CNN representations with semantic visual concepts, in order to explain CNN predictions quantitatively and semantically. We propose to learn the decision tree without strong supervision for explanations. Our method is a generic approach and has been successfully applied to various benchmark CNNs. Experiments have demonstrated the effectiveness of our method.

\section{Related work}

In this section, we limit our discussion to the literature of opening the black box of CNN representations. \cite{Interpretability,modelInterpretability,modelInterpretability2,modelInterpretability3} discussed the definition of interpretability from different perspectives with respect to different tasks. Zhang \emph{et al.}~\cite{InterpretabilitySurvey} made a survey for the interpretability of deep visual models.

\textbf{CNN visualization:}{\verb| |} Visualization of filters in a CNN is the most direct way of exploring the pattern hidden inside a neural unit. Gradient-based visualization~\cite{CNNVisualization_1,CNNVisualization_2} estimates the input image that maximizes the activation score of a neural unit. Up-convolutional nets~\cite{FeaVisual} invert feature maps of conv-layers into images. Unlike gradient-based methods, up-convolutional nets cannot mathematically ensure that the visualization result reflects actual neural representations.

Zhou \emph{et al.}~\cite{CNNSemanticDeep} proposed a method to accurately compute the image-resolution receptive field of neural activations in a feature map. The estimated receptive field of a neural activation is smaller than the theoretical receptive field based on the filter size. The accurate estimation of the receptive field is crucial to understand a filter's representations. Bau \emph{et al.}~\cite{Interpretability} further defined six types of semantics for CNNs, \emph{i.e.} objects, parts, scenes, textures, materials, and colors. Zhang \emph{et al.}~\cite{explanatoryGraph} summarized the six types of semantics into ``parts'' and ``textures.'' Nevertheless, each filter in a CNN represents a mixture of semantics. \cite{interpretableDecomposition} explained semantic reasons for visual recognition.

\textbf{Network diagnosis:}{\verb| |} Going beyond visualization, some methods diagnose a pre-trained CNN to obtain insight understanding of CNN representations.

Fong and Vedaldi~\cite{net2vector} analyzed how multiple filters jointly represented a certain semantic concept. Yosinski \emph{et al.}~\cite{CNNAnalysis_2} evaluated the transferability of filters in intermediate conv-layers. Aubry \emph{et al.}~\cite{CNNVisualization_5} computed feature distributions of different categories in the CNN feature space. Selvaraju \emph{et al.}~\cite{visualCNN_grad_2} and Fong \emph{et al.}~\cite{visualCNN_grad} propagated gradients of feature maps \emph{w.r.t.} the CNN loss back to the image, in order to estimate image regions that directly contribute the network output. The LIME~\cite{trust} and SHAP~\cite{shap} extracted image regions that were used by a CNN to predict a label. Zhang \emph{et al.}~\cite{explainer} used an explainer network to interpret object-part representations in intermediate layers of CNNs.

Network-attack methods~\cite{CNNInfluence,CNNAnalysis_1} diagnosed network representations by computing adversarial samples for a CNN. In particular, influence functions~\cite{CNNInfluence} were proposed to compute adversarial samples, provide plausible ways to create training samples to attack the learning of CNNs, fix the training set, and further debug representations of a CNN. Lakkaraju \emph{et al.}~\cite{banditUnknown} discovered knowledge blind spots (unknown patterns) of a pre-trained CNN in a weakly-supervised manner. The study of \cite{CNNBias} examined representations of conv-layers and automatically discover potential, biased representations of a CNN due to the dataset bias.

\textbf{CNN semanticization:}{\verb| |} Compared to the diagnosis of CNN representations, some studies aim to learn more meaningful CNN representations. Some studies extracted neural units with certain semantics from CNNs for different applications. Given feature maps of conv-layers, Zhou \emph{et al.}~\cite{CNNSemanticDeep} extracted scene semantics. Simon~\emph{et al.} mined objects from feature maps of conv-layers~\cite{ObjectDiscoveryCNN_2}, and learned object parts~\cite{CNNSemanticPart}. The capsule net~\cite{capsule} used a dynamic routing mechanism to parse the entire object into a parsing tree of capsules. Each output dimension of a capsule in the network may encode a specific meaning. Zhang \emph{et al.}~\cite{interpretableCNN} proposed to learn CNNs with disentangled intermediate-layer representations. The infoGAN~\cite{infoGAN} and $\beta$-VAE~\cite{betaVAE} learned interpretable input codes for generative models. Zhang \emph{et al.}~\cite{transplant} learned functionally interpretable, modular structures for neural networks via network transplanting.

\textbf{Decision trees for neural networks:}{\verb| |} Distilling knowledge from neural networks into tree structures is an emerging direction~\cite{distillDecisionTree,TreeDistill,TreeDistill2}, but the trees did not explain the network knowledge at a human-interpretable semantic level. Wu \emph{et al.}~\cite{RNNTree} learned a decision tree via knowledge distillation to represent the output feature space of an RNN, in order to regularize the RNN for better representations. Vaughan \emph{et al.}~\cite{additiveExplainer} distilled knowledge into an additive model for explanation.

In spite of the use of tree structures, there are two main differences between the above two studies and our research. Firstly, we focus on using a tree to semantically explain each prediction made by a pre-trained CNN. In contrast, decision trees in above studies are mainly learned for classification and cannot provide semantic-level explanations. Secondly, we summarize decision modes from gradients \emph{w.r.t.} neural activations of object parts as rationales to explain CNN prediction. Compared to above ``distillation-based'' methods, our ``gradient-based'' decision tree reflects CNN predictions more directly and strictly.

\begin{figure}[t]
\centering
\includegraphics[width=0.9\linewidth]{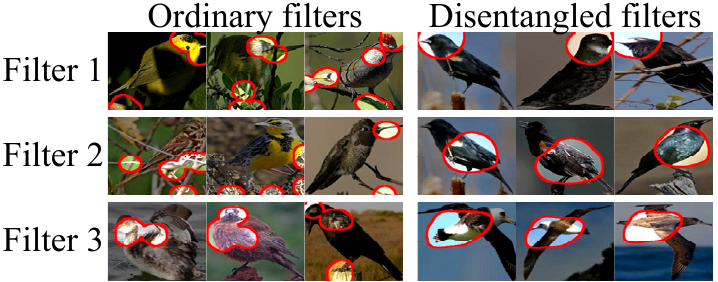}
\caption{Comparisons between ordinary CNN feature maps and disentangled feature maps that are used in this study. We visualize image regions corresponding to each feature map based on \cite{CNNSemanticDeep}.}
\label{fig:disentanglement}
\end{figure}

\section{Image-specific rationale of a CNN prediction}

In this section, we design a method to simplify the complex feature processing inside a CNN into a linear form (\emph{i.e.} Equation~(\ref{eqn:rationale})), as the specific rationale of the prediction \emph{w.r.t.} the input image. This clarifies (i) which object parts activate which filters in the CNN and (ii) how much these parts/filters contribute to the final prediction score.

In order to obtain semantic-level rationale of a CNN prediction, we need (i) first to ensure that the CNN's middle-layer features are semantically meaningful, and (ii) then to extract explicit contributions of middle-layer features to the prediction score.

In this study, we learn the CNN for object classification. Theoretically, we can interpret CNNs oriented to different tasks. Nevertheless, in this paper, we limit our attention to CNNs for classification, in order to simplify the story.

\subsection{Learning disentangled filters}

The basic idea is to revise a benchmark CNN, in order to make each filter in the top conv-layer represent a specific object part. We expect the filter to be automatically converged to the representation of a part, instead of using additional part annotations to supervise the learning process.

We apply the filter loss~\cite{interpretableCNN} to each filter in the top conv-layer to push the filter towards the representation of an object part. As shown in Fig.~\ref{fig:disentanglement}, the filter is learned to be activated by the same object part given different input images. Theoretically, our method also supports other techniques of mining interpretable features in middle layers~\cite{explanatoryGraph,ObjectDiscoveryCNN_2}. Nevertheless, the filter loss usually ensures more meaningful features than the other approaches.

\textbf{Filter loss:} Let $x_{f}\in\mathbb{R}^{L\times L}$ denote the feature map of a filter $f$. Without part annotations, the filter loss forces $x_{f}$ to be exclusively activated by a specific part of a category. We can summarize the filter loss as the minus mutual information between the distribution of feature maps and that of part locations.
\begin{small}
\begin{equation}
\begin{split}
{\bf Loss}_{f}=&{\sum}_{x_{f}\in{\bf X}_{f}}Loss_{f}(x_{f})=-MI({\bf X}_{f};{\bf P})\\
=&-{\sum}_{\mu\in{\bf P}}p(\mu){\sum}_{x_{f}\in{\bf X}_{f}}p(x_{f}|\mu)\log\frac{p(x_{f}|\mu)}{p(x_{f})}
\end{split}
\end{equation}
\end{small}
where {\small$MI(\cdot)$} indicates the mutual information. {\small${\bf X}_{f}$} denotes a set of feature maps of $f$ extracted from different input images. {\small${\bf P}=\{\mu|\mu=[h,w],1\leq h,w\leq L\}\cup\{\emptyset\}$} is referred to as a set of all part-location candidates. Each location $\mu=[h,w]$ corresponds to an activation unit in $x_{f}$. Besides, $\emptyset\in{\bf P}$ denotes the case that the target part does not appear in the input image. In this case, all units in $x_{f}$ are expected to keep inactivated. The joint probability $p(x_{f},\mu)$ to describe the compatibility between $x_{f}$ and $\mu$ (please see \cite{interpretableCNN} for details).

The filter loss ensures that given an input image, $x_{f}$ should match only one of all $L^2+1$ location candidates. It is assumed that repetitive shapes on various regions are more likely to describe low-level textures than high-level parts. If the part appears, $x_{f}$ should have a single activation peak at the part location; otherwise, $x_{f}$ should keep inactivated.

\subsection{Quantitative rationales of CNN predictions}

As analyzed in \cite{Interpretability}, filters in high conv-layers are more prone to represent object parts, while those in low conv-layers usually describe textures. Therefore, we choose filters in the top conv-layer to represent object parts. Consequently, we quantitatively analyze how fully-connected (FC) layers use object-part features from the top conv-layer to make final predictions, as the rationale.

Given an input image $I$, let {\small$x\in\mathbb{R}^{L\times L\times D}$} denote the feature map of the top conv-layer after a ReLU operation, where $L$ denotes the height/width of the feature map, and $D$ is the filter number. Let $y$ denote the scalar classification score of a certain category before the softmax operation (although the CNN is usually learned for multiple categories). Our task is to use $x$ to represent the rationale of $y$.

As discussed in \cite{shap,trust}, we can use a piecewise linear representation to represent the function of cascaded FC layers and ReLU layers, as follows.
\begin{small}
\begin{equation}
y=f_{\textrm{fc-n}}(f_{\textrm{relu}}(\cdots f_{\textrm{fc-1}}(x)))=\sum_{h,w,d}g^{(h,w,d)}\cdot x^{(h,w,d)}+b
\end{equation}
\end{small}
where {\small$x^{(h,w,d)}\in\mathbb{R}$} denotes the element at the location $(h,w)$ of the $d$-th channel; {\small$g^{(h,w,d)}$} is a weight that describes the importance of {\small$x^{(h,w,d)}$} for the prediction on $I$. Theoretically, we can compute {\small$g\!=\!\frac{\partial y}{\partial x}$} and {\small$b\!=\!y-g\otimes x$}.

We use weights $g$ to denote the specific \textit{rationale} of the prediction for the input image. {\small$g^{(h,w,d)}x^{(h,w,d)}$} measures {\small$x^{(h,w,d)}$}'s \textit{quantitative contribution} to the prediction.

Different input images correspond to different weights $g$, \emph{i.e.} different \textit{rationales} of their CNN predictions. It is because different images have various signal passing routes through ReLU layers. Given an input images $I$, the CNN uses certain weight values that are specific to $I$.

Because each interpretable filter only has a single activation peak~\cite{interpretableCNN}, we can further compute vectors {\small${\bf x},{\bf g}\in\mathbb{R}^{D}$} as an approximation to the tensors {\small$x,\frac{\partial y}{\partial x}\in\mathbb{R}^{L\times L\times D}$} to simplify the computation. We get {\small${\bf x}^{(d)}\!=\!\frac{1}{s_{d}}\sum_{h,w}x^{(h,w,d)}$} and {\small${\bf g}^{(d)}\!=\!\frac{s_{d}}{L^2}\sum_{h,w}\frac{\partial y}{\partial x^{(h,w,d)}}$}, where {\small${\bf x}^{(d)}$} denotes the $d$-th element of {\small${\bf x}$}. {\small$s_{d}\!=\!{\mathbb{E}}_{I}{\mathbb{E}}_{h,w}\,x^{(h,w,d)}$} is used to normalize the activation magnitude of the $d$-th filter.

In this way, we can consider {\small${\bf x}$} and {\small${\bf g}$} to represent prediction rationales\footnote[4]{Without loss of generality, we normalize ${\bf g}$ to a unit vector for more convincing results: {$y\!\leftarrow\!{y}/{\Vert{\bf g}\Vert}$}, {${\bf g}\!\leftarrow\!{{\bf g}}/{\Vert{\bf g}\Vert}$}, and {$b\!\leftarrow\!{b}/{\Vert{\bf g}\Vert}$}.}, \emph{i.e.} using which filters/parts for prediction.
\begin{small}
\begin{equation}
y\approx{\bf g}^{T}{\bf x}+b\label{eqn:rationale}
\end{equation}
\end{small}
Different dimensions of the vector {\small${\bf x}$} measure the scalar signal strength of different object parts, since a filter potentially represents a certain object part. {\small${\bf g}$} corresponds to the selection of object parts for the CNN prediction.

\begin{figure}[t]
\centering
\includegraphics[width=0.9\linewidth]{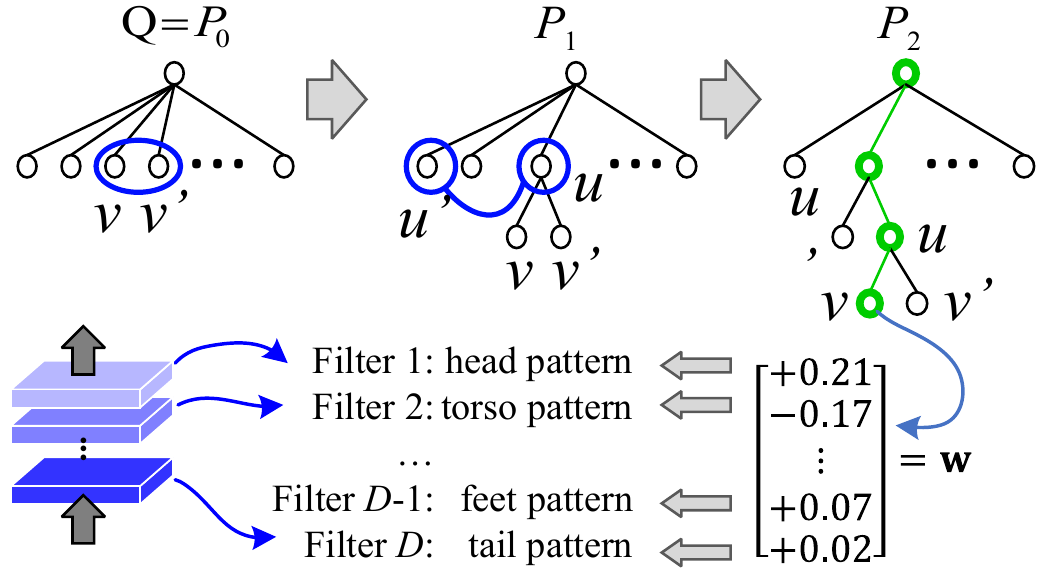}
\caption{Process of learning a decision tree. Green lines in $P_{3}$ indicate a parse tree to explain the rationale of the prediction on an image.}
\label{fig:merge}
\end{figure}

\section{Learning a decision tree}

We learn a decision tree to interpret the classification of each category. In the following two subsections, we first define basic concepts in a decision tree and then introduce the learning algorithm.

\subsection{Decision tree}

Let us focus on the decision tree for a certain category. We consider images of this category as positive images and consider other images as negative images. {\small${\boldsymbol\Omega}^{+}$} denotes image indexes of the target category, \emph{i.e.} positive images, and {\small${\boldsymbol\Omega}={\boldsymbol\Omega}^{+}\cup{\boldsymbol\Omega}^{-}$} represents all training images. For an image {\small$I_{i}$} ({\small$i\in{\boldsymbol\Omega}$}), $y_{i}$ denotes the classification score of the target category before the softmax layer.

As shown in Fig.~\ref{fig:decision}, each node $v$ in the decision tree encodes a decision mode that is hidden inside FC layers of the CNN. A decision mode represents a common rationale of the prediction shared by a group of positive training images {\small$\Omega_{v}\subset{\boldsymbol\Omega}^{+}$}. The decision tree organizes the hierarchy of decision modes in a coarse-to-fine manner from the root node to leaf nodes. Children nodes {\small$v'\in Child(v)$} divides the parent $v$'s decision mode into fine-grained modes. Fine-grained modes are shared by sub-groups of images.

\begin{algorithm}[t]
\caption{Learning a decision tree for a category}
\label{alg}
\begin{algorithmic}
\STATE {\bf Input:} 1. A CNN with disentangled filters, 2. training images {\small${\boldsymbol\Omega}={\boldsymbol\Omega}^{+}\cup{\boldsymbol\Omega}^{-}$}.
\STATE {\bf Output:} A decision tree.
\STATE Initialize a tree $Q=P_0$ and set $t=0$
\FOR{each image $I_{i}$, {\small$i\in{\boldsymbol\Omega}^{+}$}}
\STATE Initialize a child of the root of the initial tree $Q$ by setting {\small$\overline{\bf g}\!=\!{\bf g}_{i}$} based on Equation (\ref{eqn:rationale}) and {\small${\boldsymbol\alpha}\!=\!{\bf 1}$}.
\ENDFOR
\FOR{{\small$t=t+1$} {\bf until} $\Delta\log E\leq0$}
\STATE 1. Choose $(v,v')$ in the second tree layer of $P_{t-1}$ that maximize $\Delta\log E$ based on Equation~(\ref{eqn:E})
\STATE 2. Merge $(v,v')$ to generate a new node $u$ based on Equations~(\ref{eqn:learnnode}) and (\ref{eqn:learnnode2}), and obtain the tree $P_{t}$.
\ENDFOR
\STATE Assign filters with semantic object parts to obtain ${\bf A}$.
\end{algorithmic}
\end{algorithm}

Just like the rationale defined in Equation~(\ref{eqn:rationale}), the decision mode in node $v$ is parameterized with ${\bf w}$ and $b$, and the mode explains predictions on a certain set of images {\small$\Omega_{v}$}. For each image {\small$I_{i}$}, {\small$i\in\Omega_{v}$}, the decision mode is given as
\begin{small}
\begin{eqnarray}
&h_{v}({\bf x}_{i})={\bf w}^{T}{\bf x}_{i}+b,\qquad {\bf w}={\boldsymbol\alpha}\circ\overline{\bf g}\\
&\underset{\overline{\bf g}}{\max}{\sum}_{i\in\Omega_{v}}cosine({\bf g}_{i},\overline{\bf g}),\quad\emph{s.t.}\;\;\overline{\bf g}^{T}\overline{\bf g}=1\label{eqn:learnnode}\\
&\min_{{\boldsymbol\alpha},b}\frac{1}{\Vert\Omega_{v}\Vert}{\sum}_{i\in\Omega_{v}}({\bf w}^{T}{\bf x}_{i}+b-y_{i})^2+\lambda\Vert{\boldsymbol\alpha}\Vert_1\label{eqn:learnnode2}
\end{eqnarray}
\end{small}
where ${\bf w}$ is referred to as the rationale of the decision mode. {\small$\overline{\bf g}$} is a unit vector ({\small$\Vert\overline{\bf g}\Vert_2\!=\!1$}) that reflects common rationales that are shared by all images in {\small$\Omega_{v}$}. ${\boldsymbol\alpha}\in\{0,1\}^{D}$ is given as a binary selection of filters in the decision mode. $\circ$ denote element-wise multiplications. We compute sparse ${\boldsymbol\alpha}$ to obtain sparse explanations for the decision mode\footnotemark[5].

In particular, when $v$ is a leaf node, the decision mode is formulated as the rationale of a specific image $I_{i}$. \emph{I.e.} ${\boldsymbol\alpha}=[1,1,\ldots,1]^{T}$ and {\small${\bf w}={\boldsymbol\alpha}\circ{\bf g}_{i}={\bf g}_{i}$}, which is computed in Equation~(\ref{eqn:rationale}).

\subsection{Learning decision trees}

Just like hierarchical clustering, the basic idea of learning a decision tree is to summarize common generic decision modes from specific decision modes of different images. Algorithm~\ref{alg} shows the pseudo-code of the learning process. At the beginning, we initialize the decision mode {\small${\bf g}_{i}$} of each positive image $I_{i}$ as a leaf node by setting {\small$\overline{\bf g}\!=\!{\bf g}_{i}$} and {\small${\boldsymbol\alpha}\!=\!{\bf 1}$}. Thus, we build an initial tree $Q$ as shown in Fig.~\ref{fig:merge}, in which the root node takes decision modes of all positive images as children. Then, in each step, we select and merge two nodes $v,v'\in V$ in the second tree layer (\emph{i.e.} children of the root node) to obtain a new node $u$, where $V$ denotes the children set of the root. $u$ becomes a new child of the root node, and $v$ and $v'$ are re-assigned as $u$'s children. The image set of $u$ is defined as {\small$\Omega_{u}=\Omega_{v}\cup\Omega_{v'}$} and we learn {\small${\boldsymbol\alpha},b,\overline{\bf g}$} for $u$ based on Equations~(\ref{eqn:learnnode}) and (\ref{eqn:learnnode2}).

In this way, we gradually revise the initial tree $P_{0}=Q$ towards the final tree after $T$ merging operations as
\begin{small}
\begin{equation}
Q=P_{0}\rightarrow P_{1}\rightarrow P_{2}\rightarrow\cdots\rightarrow P_{T}=\hat{P}
\end{equation}
\end{small}

We formulate the objective for learning as follows.
\begin{small}
\begin{equation}
\max_{P}E,\qquad E=
\underbrace{\frac{\prod_{i\in{\boldsymbol\Omega}^{+}}P({\bf x}_{i})}{\prod_{i\in{\boldsymbol\Omega}^{+}}Q({\bf x}_{i})}}_{\textrm{Discrimination power}}\cdot
\underbrace{e^{-\beta\Vert V\Vert}}_{\textrm{Sparsity of}\atop\textrm{decision modes}}
\label{eqn:E}
\end{equation}
\end{small}
where $P({\bf x}_{i})$ denotes the likelihood of ${\bf x}_{i}$ being positive that is estimated by the tree $P$. {\small$\prod_{i}P({\bf x}_{i})$} indicates the discriminative power of $P$. $\beta$ is a scaling parameter\footnote[5]{Please see the experiment section for settings of $\beta$, $\gamma$, and $\lambda$.}. This objective penalizes the decrease of the discriminative power and encourages the system to summarize a few decision modes as generic explanations for CNN predictions. We compute the likelihood of ${\bf x}_{i}$ being positive as
\begin{small}
\begin{equation}
P({\bf x}_{i})={e^{\gamma\hat{h}({\bf x}_{i})}}/{{\sum}_{j\in{\boldsymbol\Omega}}e^{\gamma\hat{h}({\bf x}_{j})}}
\end{equation}
\end{small}
where {\small$\hat{h}({\bf x}_{i})\!=\!h_{\hat{v}}({\bf x}_{i})$} denotes the prediction on {\small${\bf x}_{i}$} based on best child $\hat{v}\in V$ in the second tree layer. $\gamma$ is a constant scaling parameter\footnotemark[5].

In the $t$-th step, we merge two nodes $v,v'\in V$ in the second tree layer of $P_{t-1}$ to get a new node $u$, thereby obtaining a new tree $P_{t}$. We can easily compute {\small$\Delta\log E$} \emph{w.r.t.} each pair of $(v,v')$ based on Equation~(\ref{eqn:E}). Thus, we learn the decision tree via a greedy strategy. In each step, we select and merge the nodes $v,v'\in V$ that maximize {\small$\frac{\Delta\log E}{\Vert\Omega_{v}\Vert+\Vert\Omega_{v'}\Vert}$}. We normalize {\small$\Delta\log E$} for reasonable clustering performance.

\subsection{Interpreting CNNs}

Given a testing image $I_{i}$, the CNN makes a prediction $y_{i}$. The decision tree estimates quantitative decision modes of the prediction at different fine-grained levels. During the inference procedure, we can infer a parse tree, which starts from the root node, in a top-down manner. Green lines in Fig.~\ref{fig:merge} show a parse tree. When we select the decision mode in the node $u$ as the rationale, we can further select its child $\hat{v}$ that maximizes the compatibility with the most specific rationale {\small${\bf g}_{i}$} as a more fine-grained mode:
\begin{small}
\begin{equation}
\hat{v}={\arg\!\max}_{v\in Child(u)}cosine({\bf g}_{i},{\bf w}_{v})
\end{equation}
\end{small}
where we add the subscript $v$ to differentiate the parameter of $v$ from parameters of other nodes.

A node $v$ in the parse tree provides the rationale of the prediction on image $I_{i}$ at a certain fine-grained level. We compute the vector ${\boldsymbol\rho}_{i}$ and ${\boldsymbol\varrho}_{i}$ to evaluate the contribution of different filters and that of different object parts.
\begin{small}
\begin{equation}
{\boldsymbol\rho}_{i}={\bf w}_{v}\circ{\bf x}_{i},\qquad{\boldsymbol\varrho}_{i}={\bf A}{\boldsymbol\rho}_{i}
\label{eqn:contri}
\end{equation}
\end{small}
where the $d$-th element of {\small${\boldsymbol\rho}_{i}\in\mathbb{R}^{D}$}, {\small$\rho_{i}^{(d)}$}, denotes the contribution to the CNN prediction that is made by the $d$-th filter. If {\small$\rho_{i}^{(d)}>0$}, then the $d$-th object part makes a positive contribution. If {\small$\rho_{i}^{(d)}<0$}, then the $d$-th filter makes a negative contribution. Based on visualization results in Figs.~\ref{fig:disentanglement} and \ref{fig:visualization}, we label a matrix {\small${\bf A}\in\{0,1\}^{M\times D}$} to assign each filter in the top conv-layer with a specific object part, where $M$ is the part number. Each filter is assigned to a certain part, and the annotation cost is {\small$O(M)$}. Similarly, the $m$-th element of {\small${\boldsymbol\varrho}_{i}\in\mathbb{R}^{M}$}, {\small$\varrho_{i}^{(m)}$} measures the contribution of the $m$-th part.

\section{Experiments}

\textbf{Implementation details:} We learned four types of disentangled CNNs based on structures of four benchmark CNNs, including the AlexNet~\cite{CNNImageNet}, the VGG-M network~\cite{VGG}, the VGG-S network~\cite{VGG}, the VGG-16 network~\cite{VGG}. Note that as discussed in \cite{interpretableCNN}, the filter loss in the explainer is not compatible with skip connections in residual networks~\cite{ResNet}. We followed the technique of \cite{interpretableCNN} to modify an ordinary CNN to a disentangled CNN, which changed the top conv-layer of the CNN to a disentangled conv-layer and further added a disentangled conv-layer on the top conv-layer. We used feature maps of the new top conv-layer as the input of our decision tree. We loaded parameters of all old conv-layers directly from the CNN that was pre-trained using images in the ImageNet ILSVRC 2012 dataset~\cite{ImageNet} with a loss for 1000-category classification. We initialized parameters of the new top conv-layer and all FC layers. Inspired by previous studies of \cite{interpretableCNN}, we can weaken the problem of multi-category classification as a number of single-category classification to simplify the evaluation of interpretability. Thus, we fine-tuned the CNN for binary classification of a single category from random images with the log logistic loss using three benchmark datasets. We simply set parameters as {\small$\beta\!=\!1$}, {\small$\gamma\!=\!1/{\mathbb{E}}_{i\in{\bf\Omega}^{+}}[y_{i}]$}, and {\small$\lambda\!=\!10^{-6}\sqrt{\Vert\Omega_{v}\Vert}$} in all experiments for fair comparisons.

\begin{figure*}
\centering
\includegraphics[width=0.95\linewidth]{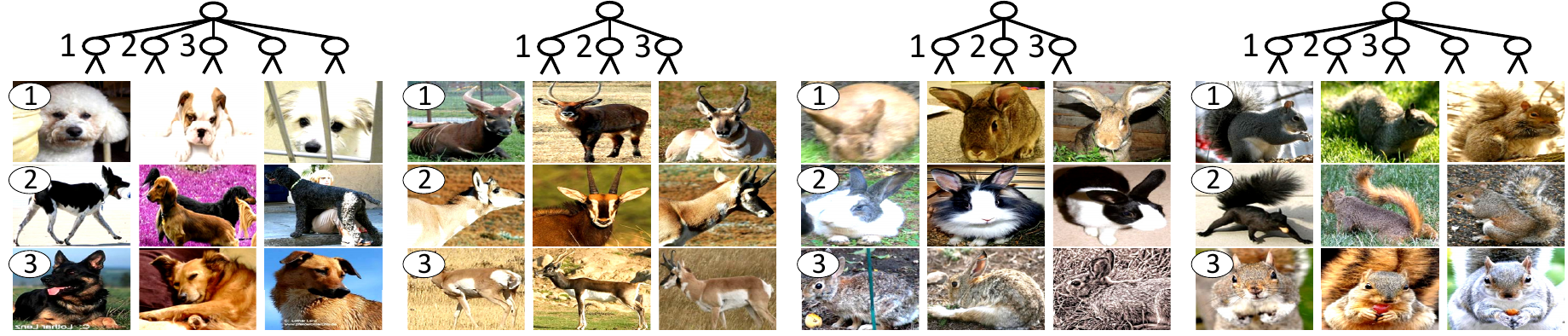}
\caption{Visualization of decision modes corresponding to nodes in the 2nd tree layer. We show typical images of each decision mode.}
\label{fig:mode}
\end{figure*}

\begin{figure*}
\centering
\includegraphics[width=0.95\linewidth]{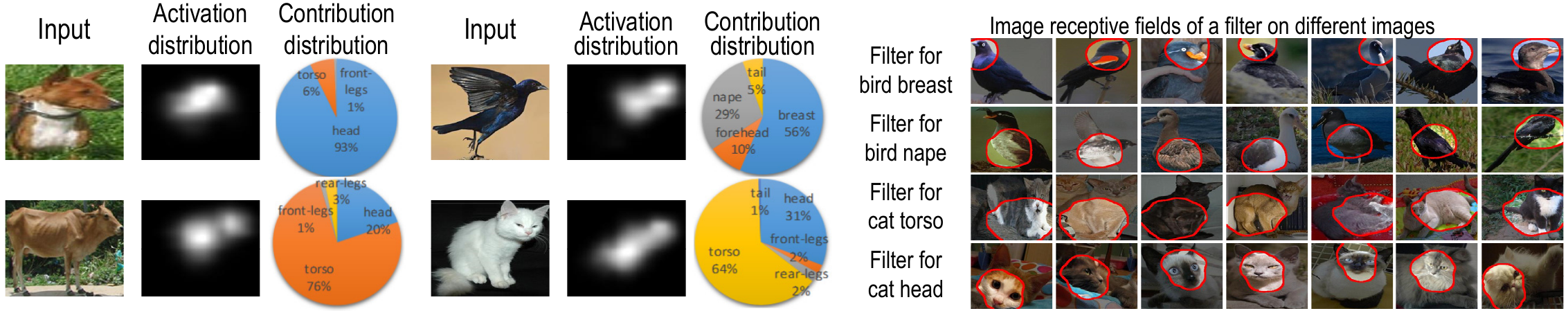}
\caption{Object-part contributions for CNN prediction. Pie charts show contribution proportions of different parts, which are estimated using nodes in the second tree layer. Heat maps indicate spatial distributions of neural activations in the top conv-layer (note that the heat maps do not represent distributions of ``contributions,'' because neural activations are not weighted by ${\bf g}_{i}$). Right figures show image receptive fields of different filters estimated by {\protect\cite{CNNSemanticDeep}}. Based on these receptive filters, we assign the filters with different object parts to compute the distribution of object-part contributions.}
\label{fig:visualization}
\end{figure*}

\textbf{Datasets:} Because the quantitative explanation of CNN predictions requires us to assign each filter in the top conv-layer with a specific object part, we used three benchmark datasets with ground-truth art annotations to evaluate our method. The selected datasets include the PASCAL-Part Dataset~\cite{SemanticPart}, the CUB200-2011 dataset~\cite{CUB200}, and the ILSVRC 2013 DET Animal-Part dataset~\cite{explanatoryGraph}. Just like in most part-localization studies~\cite{SemanticPart,explanatoryGraph}, we used animal categories, which prevalently contain non-rigid shape deformation, for evaluation. \emph{I.e.} we selected six animal categories---\textit{bird, cat, cow, dog, horse}, and \textit{sheep}---from the PASCAL Part Dataset. The CUB200-2011 dataset contains 11.8K images of 200 bird species. Like in \cite{ActivePart,CNNSemanticPart}, we ignored species labels and regarded all these images as a single bird category. The ILSVRC 2013 DET Animal-Part dataset~\cite{explanatoryGraph} consists of 30 animal categories among all the 200 categories for object detection in the ILSVRC 2013 DET dataset~\cite{ImageNet}.

\textbf{Analysis of object parts for prediction:} We analyzed the contribution of different object parts in the CNN prediction, when we assigned each filter with a specific object part. The vector ${\boldsymbol\varrho}_{i}$ in Equation~(\ref{eqn:contri}) specifies the contribution of different object parts in the prediction of $y_{i}$. For the $m$-th object part, we computed {\small$contri_{m}\!=\!\vert\varrho_{i}^{(m)}\vert/{\sum_{m'=1}^{M}\vert\varrho_{i}^{(m')}\vert}$} as the ratio of the part's contribution.

More specifically, for CNNs based on the ILSVRC 2013 DET Animal-Part dataset, we manually labeled the object part for each filter in the top conv-layer. For CNNs based on the Pascal VOC Part dataset~\cite{SemanticPart}, the study of \cite{interpretableCNN} merged tens of small parts into several major landmark parts for the six animal categories. Given a CNN for a certain category, we used \cite{CNNSemanticDeep} to estimate regions in different images that corresponded to each filter's neural activations, namely the \textit{image receptive field} of the filter (please see Figs.~\ref{fig:visualization} and \ref{fig:disentanglement}). For each filter, we selected a part from all major landmark parts, which was closest to the filter's image receptive field through all positive images. For the CNN based on the CUB200-2011 dataset, we used ground-truth positions of the breast, forehead, nape, tail of birds as major landmark parts. Similarly, we assigned each filter in the top conv-layer with the nearest landmark part.

\textbf{Evaluation metrics:} The evaluation has two aspects. Firstly, we use two metrics to evaluate the accuracy of the estimated rationale of a prediction. The first metric evaluates \textit{errors of object-part contributions} to the CNN prediction that were estimated using nodes in the second tree layer. Given an input image $I$, ${\boldsymbol\varrho}_{i}$ in Equation~(\ref{eqn:contri}) denotes the quantitative contribution of the $i$-th part. Accordingly, {\small${\boldsymbol\varrho}_{i}^{*}\!=\!y-\hat{y}_{i}$} is referred to as the ground-truth contribution of the part, where $y$ denotes the original CNN prediction on $I$; $\hat{y}_{i}$ is the output when we removed neural activations from feature maps (filters) corresponding to the $i$-th part. In this way, we used the deviation $\mathbb{E}_{I\in{\bf I}}[{\boldsymbol\varrho}_{i}-{\boldsymbol\varrho}_{i}^{*}]/\mathbb{E}_{I\in{\bf I}}[y]$ to denote the error of the $i$-th part contributions. Another metric, namely the \textit{fitness of contribution distributions}, compares the ground-truth contribution distribution over different filters in the top conv-layer with the estimated contribution of these filters during the prediction process. When the decision tree uses node $\hat{v}$ to explain the prediction for $I_{i}$, the vector ${\boldsymbol\rho}_{i}$ in Equation~(\ref{eqn:contri}) denotes the estimated contribution distribution of different filters. {\small${\bf t}_{i}\!=\!{\bf g}_{i}\circ{\bf x}_{i}$} corresponds to the ground-truth contribution distribution. We reported the interaction-of-the-union value between ${\boldsymbol\rho}_{i}$ and ${\bf t}_{i}$ to measure the fitness of the ground-truth and the estimated filter contribution distributions. \emph{I.e.} we computed the fitness as {\small${\mathbb{E}}_{i\in{\bf\Omega}^{+}}[\frac{\min(\hat{\rho}_{i}^{(d)},\vert t_{i}^{(d)}\vert)}{\max(\hat{\rho}_{i}^{(d)},\vert t_{i}^{(d)}\vert)}]$}, where $t_{i}^{(d)}$ denotes the $d$-th element of ${\bf t}_{i}$ and {\small$\hat{\rho}_{i}^{(d)}\!=\!\max\{\rho_{i}^{(d)}sign(t_{i}^{(d)}),0\}$}. We used non-negative values of $\hat{\rho}_{i}^{(d)}$ and $\vert t_{i}^{(d)}\vert$, because vectors ${\boldsymbol\rho}_{i}$ and ${\bf t}_{i}$ may have negative elements.

Secondly, in addition to the accuracy of the rationale, we also measured the information loss of using the decision tree to represent a CNN, as a supplementary evaluation. A metric is the \textit{classification accuracy}. Because $\hat{h}({\bf x}_{i})$ denotes the prediction of $y_{i}$ based on the best child in the second tree layer, we regarded $\hat{h}(\cdot)$ as the output of the tree and we evaluated the discrimination power of $\hat{h}(\cdot)$. We used values of $\hat{h}({\bf x}_{i})$ for classification and compared its classification accuracy with the accuracy of the CNN. Another metric, namely the \textit{prediction error}, measures the error of the estimated value $\hat{h}({\bf x}_{i})$ \emph{w.r.t} the true value $y_{i}$. We computed the prediction error as {\small${\mathbb{E}}_{i\in{\bf\Omega}^{+}}[\vert\hat{h}({\bf x}_{i})-y_{i}\vert]/(\max_{i\in{\bf\Omega}}y_{i}-\min_{i\in{\bf\Omega}}y_{i})$}, where we normalized the error using the value range of $y_{i}$.

Evaluation for nodes in different layers: The above three metrics evaluate decision modes (nodes) in the second layer of the decision tree. Because nodes in lower layers encode more fine-grained decision modes, we extended the three metrics to evaluate nodes in low layers. When we evaluated nodes in the $k$-th layer, we temporarily constructed a new tree by removing all nodes above the $k$-th layer and directly connecting the root node to nodes in the $k$-th layer. Thus, we can apply the evaluation to the new tree.

\begin{table}[t]
\centering
\resizebox{1.0\linewidth}{!}{\begin{tabular}{c|ccccc}
\hline
Dataset & 2nd & 5th & 10th & 50th & 100th\\
\hline
ILSVRC Animal-Part & 4.8 & 31.6 & 69.1 & 236.5 & 402.1\\
VOC Part & 3.8 & 25.7 & 59.0 & 219.5 & 361.5\\
CUB200-2011 & 5.0 & 32.0 & 64.0 & 230.0 & 430.0\\
\hline
\end{tabular}}
\caption{Average number of nodes in the 2nd, 5th, 10th, 50th, and 100th layer of decision trees learned for VGG-16 nets.}
\label{tab:size}
\end{table}

\begin{table}[t]
\centering
\resizebox{0.9\linewidth}{!}{\begin{tabular}{c|cccc|c}
&\!\! breast \!\!&\!\! forehead \!\!&\!\! nape \!\!&\!\! tail \!\!&\!\! average\\
\hline
2nd layer & 0.028 & 0.004 & 0.013 & 0.005 & 0.013\\
5th layer & 0.024 & 0.004 & 0.010 & 0.006 & 0.011\\
10th layer & 0.022 & 0.004 & 0.010 & 0.005 & 0.010\\
50th layer & 0.018 & 0.003 & 0.008 & 0.005 & 0.009\\
100th layer & 0.019 & 0.003 & 0.008 & 0.005 & 0.009\\
\end{tabular}}
\vspace{2pt}
\caption{Errors of object-part contributions based on nodes in the 2nd/5th/10th/50th/100th layer of the decision tree. We use the error of object-part contributions to evaluate the estimated rationale of a prediction. The CNN was learned using bird images of the CUB200 dataset.}
\label{tab:contri}
\end{table}

\begin{table}[t]
\centering
\resizebox{1.0\linewidth}{!}{\begin{tabular}{c|c|cccccc}
\hline
\multicolumn{2}{c|}{Dataset} &\!\! 2nd \!\!&\!\! 5th \!\!&\!\! 10th \!\!&\!\! 50th \!\!&\!\! 100th \!\!&\!\! leaves\!\!\\
\hline
\multirow{3}{*}{{\footnotesize VGG-16\;}} & ILSVRC Animal-Part & 0.23 & 0.30 & 0.36 & 0.52 & 0.65 & 1.00\\
& VOC Part & 0.22 & 0.30 & 0.36 & 0.53 & 0.67 & 1.00\\
& CUB200-2011 & 0.21 & 0.26 & 0.28 & 0.33 & 0.37 & 1.00\\
\hline
\multirow{2}{*}{{\footnotesize VGG-M\;}} & VOC Part
& 0.35 & 0.38 & 0.46 & 0.63 & 0.78 & 1.00\\
& CUB200-2011
& 0.44 & 0.44 & 0.46 & 0.59 & 0.63 & 1.00\\
\hline
\multirow{2}{*}{{\footnotesize VGG-S\;}} & VOC Part
& 0.33 & 0.35 & 0.41 & 0.63 & 0.80 & 1.00\\
& CUB200-2011
& 0.40 & 0.40 & 0.43 & 0.48 & 0.52 & 1.00\\
\hline
\multirow{2}{*}{{\footnotesize AlexNet\;}} & VOC Part
& 0.37 & 0.38 & 0.47 & 0.66 & 0.82 & 1.00\\
& CUB200-2011
& 0.47 & 0.47 & 0.47 & 0.58 & 0.66 & 1.00\\
\hline
\end{tabular}}
\vspace{0pt}
\caption{Average fitness of contribution distributions based on nodes in the 2nd/5th/10th/50th/100th layer and leaf nodes. We use the fitness of contribution distributions to evaluate the accuracy of the estimated rationale of a prediction.}
\label{tab:IOU}
\end{table}

\begin{table}[t]
\centering
\resizebox{1.0\linewidth}{!}{
\begin{tabular}{c|c|ccccccc}
\multicolumn{9}{c}{Average classification accuracy}\\
\hline
\multicolumn{2}{c|}{Dataset} &\!\! CNN \!\!&\!\! 2nd \!\!&\!\! 5th \!\!&\!\! 10th \!\!&\!\! 50th \!\!&\!\! 100th \!\!&\!\! leaves\!\!\\
\hline
\multirow{3}{*}{{\footnotesize VGG-16}} &\!\!ILSVRC Animal-Part \!\!&\!\! 96.7 \!\!&\!\! 94.4 \!\!&\!\! 89.0 \!\!&\!\! 88.7 \!\!&\!\! 88.6 \!\!&\!\! 88.7 \!\!&\!\! 87.8\!\!\\
&\!\!VOC Part \!\!&\!\! 95.4 \!\!&\!\! 94.2 \!\!&\!\! 91.0 \!\!&\!\! 90.1 \!\!&\!\! 89.8 \!\!&\!\! 89.4 \!\!&\!\! 88.2\!\!\\
&\!\!CUB200-2011 \!\!&\!\! 96.5 \!\!&\!\! 91.5 \!\!&\!\! 92.2 \!\!&\!\! 88.3 \!\!&\!\! 88.6 \!\!&\!\! 88.9 \!\!&\!\! 85.3\!\!\\
\hline
\multirow{2}{*}{{\footnotesize VGG-M}} &\!\!VOC Part
\!\!&\!\!94.2
\!\!&\!\!95.7
\!\!&\!\!94.2
\!\!&\!\!93.1
\!\!&\!\!93.0
\!\!&\!\!92.6
\!\!&\!\!90.8
\\
&\!\!CUB200-2011
\!\!&\!\!96.0			
\!\!&\!\!97.2
\!\!&\!\!96.8
\!\!&\!\!96.0
\!\!&\!\!95.2
\!\!&\!\!94.9
\!\!&\!\!93.5
\\
\hline
\multirow{2}{*}{{\footnotesize VGG-S}} &\!\!VOC Part
\!\!&\!\!95.5
\!\!&\!\!92.7
\!\!&\!\!92.6
\!\!&\!\!91.3
\!\!&\!\!90.2
\!\!&\!\!88.8
\!\!&\!\!86.1
\\
&\!\!CUB200-2011
\!\!&\!\!95.8
\!\!&\!\!95.4
\!\!&\!\!94.9
\!\!&\!\!93.1
\!\!&\!\!93.4
\!\!&\!\!93.6
\!\!&\!\!88.8
\\
\hline
\multirow{2}{*}{{\footnotesize AlexNet}} &\!\!VOC Part
\!\!&\!\!93.9
\!\!&\!\!90.7
\!\!&\!\!88.6
\!\!&\!\!88.6
\!\!&\!\!87.9
\!\!&\!\!86.2
\!\!&\!\!84.1
\\
&\!\!CUB200-2011
\!\!&\!\!95.4
\!\!&\!\!94.9
\!\!&\!\!94.2
\!\!&\!\!94.3
\!\!&\!\!92.8
\!\!&\!\!92.0
\!\!&\!\!90.0
\\
\hline
\end{tabular}}
\resizebox{1.0\linewidth}{!}{
\begin{tabular}{c|c|cccccc}
\multicolumn{7}{c}{Average prediction error} & \textcolor{white}{\Huge A}\\
\hline
\multicolumn{2}{c|}{Dataset} &\!\! 2nd \!\!&\!\! 5th \!\!&\!\! 10th \!\!&\!\! 50th \!\!&\!\! 100th \!\!&\!\! leaves\!\!\\
\hline
\multirow{3}{*}{{\footnotesize VGG-16}} &\!\!ILSVRC Animal-Part \!\!&\!\! 0.052 \!\!&\!\! 0.064 \!\!&\!\! 0.063 \!\!&\!\! 0.049 \!\!&\!\! 0.034 \!\!&\!\! 0.00\\
&\!\!VOC Part \!\!&\!\! 0.052 \!\!&\!\! 0.066 \!\!&\!\! 0.070 \!\!&\!\! 0.051 \!\!&\!\! 0.035 \!\!&\!\! 0.00\\
&\!\!CUB200-2011 \!\!&\!\! 0.075 \!\!&\!\! 0.099 \!\!&\!\! 0.101 \!\!&\!\! 0.087 \!\!&\!\! 0.083 \!\!&\!\! 0.00\\
\hline
\multirow{2}{*}{{\footnotesize VGG-M}} &\!\!VOC Part
\!\!&\!\!0.053
\!\!&\!\!0.051
\!\!&\!\!0.051
\!\!&\!\!0.034
\!\!&\!\!0.019
\!\!&\!\! 0.00\\
&\!\!CUB200-2011
\!\!&\!\!0.036
\!\!&\!\!0.037
\!\!&\!\!0.038
\!\!&\!\!0.035
\!\!&\!\!0.030
\!\!&\!\! 0.00\\
\hline
\multirow{2}{*}{{\footnotesize VGG-S}} &\!\!VOC Part
\!\!&\!\!0.047
\!\!&\!\!0.047
\!\!&\!\!0.045
\!\!&\!\!0.035
\!\!&\!\!0.019
\!\!&\!\! 0.00\\
&\!\!CUB200-2011
\!\!&\!\!0.045
\!\!&\!\!0.046
\!\!&\!\!0.050
\!\!&\!\!0.051
\!\!&\!\!0.038
\!\!&\!\! 0.00\\
\hline
\multirow{2}{*}{{\footnotesize AlexNet}} &\!\!VOC Part
\!\!&\!\!0.055
\!\!&\!\!0.058
\!\!&\!\!0.055
\!\!&\!\!0.038
\!\!&\!\!0.020
\!\!&\!\! 0.00\\
&\!\!CUB200-2011
\!\!&\!\!0.044
\!\!&\!\!0.044
\!\!&\!\!0.045
\!\!&\!\!0.039
\!\!&\!\!0.033
\!\!&\!\! 0.00\\
\hline
\end{tabular}}
\vspace{0pt}
\caption{Average classification accuracy and average prediction error based on nodes in the 2nd/5th/10th/50th/100th layer and leaf nodes of the tree. We use the classification accuracy and the prediction error to measure the information loss when using a decision tree to represent a CNN.}
\label{tab:ClassificationAndError}
\end{table}

\textbf{Explanations based on the decision tree:} Decision modes in the learned decision tree objectively reflected the knowledge hidden inside a CNN. Table~\ref{tab:size} shows the structure of the decision tree by listing numbers of nodes in different layers of the decision tree. Fig.~\ref{fig:mode} visualizes decision modes in the decision tree. Fig.~\ref{fig:visualization} shows distributions of object-part contributions to the CNN prediction, which were estimated using nodes in the second layer of decision trees.

Table~\ref{tab:ClassificationAndError} evaluates the information loss when we use the decision tree to represent a CNN. Metrics of the average classification accuracy, the average prediction error are used for evaluation. Tables~\ref{tab:contri} and \ref{tab:IOU} use errors of object-part contributions and the average fitness of contribution distributions, respectively, to evaluate the accuracy of the estimated rationales based on nodes in different tree layers. Generally speaking, because fine-grained decision modes are close to the image-specific rationale, fine-grained decision modes usually yielded lower error prediction rates. However, fine-grained decision modes did not exhibit higher accuracy in classification. It is because our method is designed to mine common decision modes for objects of a certain category, and ignores random/negative images, which is different from the discriminative learning of classifiers.

\section{Conclusion and discussions}

In this study, we use a decision tree to explain CNN predictions at the semantic level. We have developed a method to revise a CNN and built a tight coupling of the CNN and a decision tree. The proposed decision tree encodes decision modes of the CNN as quantitative rationales for each CNN prediction. Our method does not need any annotations of object parts or textures in training images to guide the learning the CNN. We have tested our method in different benchmark datasets, and experiments have proved the effectiveness of our approach.

Note that theoretically, the decision tree just provides an approximate explanation for CNN predictions, instead of an accurate reconstruction of CNN representation details. There are two reasons. Firstly, without accurate object-part annotations to supervised the learning of CNNs, the filter loss can only roughly make each filter to represent an object part. The filter may produce incorrect activations in a few challenging images. Secondly, the decision mode in each node ignores insignificant object-part filters to ensure a sparse representation of the decision mode.

{\small
\bibliographystyle{ieee}
\bibliography{TheBib}

\begin{thebibliography}{10}\itemsep=-1pt

\bibitem{CNNVisualization_5}
M.~Aubry and B.~C. Russell.
\newblock Understanding deep features with computer-generated imagery.
\newblock {\em In {ICCV}}, 2015.

\bibitem{Interpretability}
D.~Bau, B.~Zhou, A.~Khosla, A.~Oliva, and A.~Torralba.
\newblock Network dissection: Quantifying interpretability of deep visual
  representations.
\newblock {\em In CVPR}, 2017.

\bibitem{ActivePart}
S.~Branson, P.~Perona, and S.~Belongie.
\newblock Strong supervision from weak annotation: Interactive training of
  deformable part models.
\newblock {\em In {ICCV}}, 2011.

\bibitem{modelInterpretability3}
A.~Chandrasekaran, V.~Prabhu, D.~Yadav, P.~Chattopadhyay, and D.~Parikh.
\newblock Do explanations make vqa models more predictable to a human?
\newblock {\em In EMNLP}, 2018.

\bibitem{TreeDistill2}
Z.~Che, S.~Purushotham, R.~Khemani, and Y.~Liu.
\newblock Interpretable deep models for icu outcome prediction.
\newblock {\em In American Medical Informatics Association (AMIA) Annual
  Symposium}, 2016.

\bibitem{infoGAN}
X.~Chen, Y.~Duan, R.~Houthooft, J.~Schulman, I.~Sutskever, and P.~Abbeel.
\newblock Infogan: Interpretable representation learning by information
  maximizing generative adversarial nets.
\newblock {\em In NIPS}, 2016.

\bibitem{SemanticPart}
X.~Chen, R.~Mottaghi, X.~Liu, S.~Fidler, R.~Urtasun, and A.~Yuille.
\newblock Detect what you can: Detecting and representing objects using
  holistic models and body parts.
\newblock {\em In {CVPR}}, 2014.

\bibitem{ImageNet}
J.~Deng, W.~Dong, R.~Socher, L.-J. Li, K.~Li, and L.~Fei-Fei.
\newblock Imagenet: A large-scale hierarchical image database.
\newblock {\em In {CVPR}}, 2009.

\bibitem{modelInterpretability2}
F.~Doshi-Velez and B.~Kim.
\newblock Towards a rigorous science of interpretable machine learning.
\newblock {\em In arXiv:1702.08608}, 2017.

\bibitem{FeaVisual}
A.~Dosovitskiy and T.~Brox.
\newblock Inverting visual representations with convolutional networks.
\newblock {\em In {CVPR}}, 2016.

\bibitem{net2vector}
R.~Fong and A.~Vedaldi.
\newblock Net2vec: Quantifying and explaining how concepts are encoded by
  filters in deep neural networks.
\newblock {\em In CVPR}, 2018.

\bibitem{visualCNN_grad}
R.~C. Fong and A.~Vedaldi.
\newblock Interpretable explanations of black boxes by meaningful perturbation.
\newblock {\em In arXiv:1704.03296v1}, 2017.

\bibitem{distillDecisionTree}
N.~Frosst and G.~Hinton.
\newblock Distilling a neural network into a soft decision tree.
\newblock {\em In arXiv:1711.09784}, 2017.

\bibitem{ResNet}
K.~He, X.~Zhang, S.~Ren, and J.~Sun.
\newblock Deep residual learning for image recognition.
\newblock {\em In {CVPR}}, 2016.

\bibitem{betaVAE}
I.~Higgins, L.~Matthey, A.~Pal, C.~Burgess, X.~Glorot, M.~Botvinick,
  S.~Mohamed, and A.~Lerchner.
\newblock $\beta$-vae: learning basic visual concepts with a constrained
  variational framework.
\newblock {\em In ICLR}, 2017.

\bibitem{patternNet}
P.-J. Kindermans, K.~T. Sch\"{u}tt, M.~Alber, K.-R. M\"{u}ller, D.~Erhan,
  B.~Kim, and S.~D\"{a}hne.
\newblock Learning how to explain neural networks: Patternnet and
  patternattribution.
\newblock {\em In ICLR}, 2018.

\bibitem{CNNInfluence}
P.~Koh and P.~Liang.
\newblock Understanding black-box predictions via influence functions.
\newblock {\em In ICML}, 2017.

\bibitem{CNNImageNet}
A.~Krizhevsky, I.~Sutskever, and G.~E. Hinton.
\newblock Imagenet classification with deep convolutional neural networks.
\newblock {\em In {NIPS}}, 2012.

\bibitem{banditUnknown}
H.~Lakkaraju, E.~Kamar, R.~Caruana, and E.~Horvitz.
\newblock Identifying unknown unknowns in the open world: Representations and
  policies for guided exploration.
\newblock {\em In AAAI}, 2017.

\bibitem{CNN}
Y.~LeCun, L.~Bottou, Y.~Bengio, and P.~Haffner.
\newblock Gradient-based learning applied to document recognition.
\newblock {\em In {Proceedings of the IEEE}}, 1998.

\bibitem{modelInterpretability}
Z.~C. Lipton.
\newblock The mythos of model interpretability.
\newblock {\em In Communications of the ACM}, 61:36--43, 2018.

\bibitem{shap}
S.~M. Lundberg and S.-I. Lee.
\newblock A unified approach to interpreting model predictions.
\newblock {\em In NIPS}, 2017.

\bibitem{CNNVisualization_2}
A.~Mahendran and A.~Vedaldi.
\newblock Understanding deep image representations by inverting them.
\newblock {\em In {CVPR}}, 2015.

\bibitem{trust}
M.~T. Ribeiro, S.~Singh, and C.~Guestrin.
\newblock ``why should i trust you?'' explaining the predictions of any
  classifier.
\newblock {\em In KDD}, 2016.

\bibitem{capsule}
S.~Sabour, N.~Frosst, and G.~E. Hinton.
\newblock Dynamic routing between capsules.
\newblock {\em In NIPS}, 2017.

\bibitem{visualCNN_grad_2}
R.~R. Selvaraju, M.~Cogswell, A.~Das, R.~Vedantam, D.~Parikh, and D.~Batra.
\newblock Grad-cam: Visual explanations from deep networks via gradient-based
  localization.
\newblock {\em In arXiv:1610.02391v3}, 2017.

\bibitem{ObjectDiscoveryCNN_2}
M.~Simon and E.~Rodner.
\newblock Neural activation constellations: Unsupervised part model discovery
  with convolutional networks.
\newblock {\em In {ICCV}}, 2015.

\bibitem{CNNSemanticPart}
M.~Simon, E.~Rodner, and J.~Denzler.
\newblock Part detector discovery in deep convolutional neural networks.
\newblock {\em In {ACCV}}, 2014.

\bibitem{VGG}
K.~Simonyan and A.~Zisserman.
\newblock Very deep convolutional networks for large-scale image recognition.
\newblock {\em In {ICLR}}, 2015.

\bibitem{CNNAnalysis_1}
C.~Szegedy, W.~Zaremba, I.~Sutskever, J.~Bruna, D.~Erhan, I.~Goodfellow, and
  R.~Fergus.
\newblock Intriguing properties of neural networks.
\newblock {\em In {arXiv:1312.6199v4}}, 2014.

\bibitem{TreeDistill}
S.~Tan, R.~Caruana, G.~Hooker, and A.~Gordo.
\newblock Transparent model distillation.
\newblock {\em In arXiv:1801.08640}, 2018.

\bibitem{additiveExplainer}
J.~Vaughan, A.~Sudjianto, E.~Brahimi, J.~Chen, and V.~N. Nair.
\newblock Explainable neural networks based on additive index models.
\newblock {\em In arXiv:1806.01933}, 2018.

\bibitem{CUB200}
C.~Wah, S.~Branson, P.~Welinder, P.~Perona, and S.~Belongie.
\newblock The caltech-ucsd birds-200-2011 dataset.
\newblock Technical report, In {California Institute of Technology}, 2011.

\bibitem{InformationBottleneck}
N.~Wolchover.
\newblock New theory cracks open the black box of deep learning.
\newblock {\em In Quanta Magazine}, 2017.

\bibitem{RNNTree}
M.~Wu, M.~C. Hughes, S.~Parbhoo, M.~Zazzi, V.~Roth, and F.~Doshi-Velez.
\newblock Beyond sparsity: Tree regularization of deep models for
  interpretability.
\newblock {\em In NIPS TIML Workshop}, 2017.

\bibitem{CNNAnalysis_2}
J.~Yosinski, J.~Clune, Y.~Bengio, and H.~Lipson.
\newblock How transferable are features in deep neural networks?
\newblock {\em In {NIPS}}, 2014.

\bibitem{CNNVisualization_1}
M.~D. Zeiler and R.~Fergus.
\newblock Visualizing and understanding convolutional networks.
\newblock {\em In {ECCV}}, 2014.

\bibitem{explanatoryGraph}
Q.~Zhang, R.~Cao, F.~Shi, Y.~Wu, and S.-C. Zhu.
\newblock Interpreting cnn knowledge via an explanatory graph.
\newblock {\em In AAAI}, 2018.

\bibitem{CNNBias}
Q.~Zhang, W.~Wang, and S.-C. Zhu.
\newblock Examining cnn representations with respect to dataset bias.
\newblock {\em In AAAI}, 2018.

\bibitem{interpretableCNN}
Q.~Zhang, Y.~N. Wu, and S.-C. Zhu.
\newblock Interpretable convolutional neural networks.
\newblock {\em In CVPR}, 2018.

\bibitem{explainer}
Q.~Zhang, Y.~Yang, Y.~Liu, Y.~N. Wu, and S.-C. Zhu.
\newblock Unsupervised learning of neural networks to explain neural networks.
\newblock {\em in arXiv:1805.07468}, 2018.

\bibitem{transplant}
Q.~Zhang, Y.~Yang, Q.~Yu, and Y.~N. Wu.
\newblock Network transplanting.
\newblock {\em in arXiv:1804.10272}, 2018.

\bibitem{InterpretabilitySurvey}
Q.~Zhang and S.-C. Zhu.
\newblock Visual interpretability for deep learning: a survey.
\newblock {\em in Frontiers of Information Technology \& Electronic
  Engineering}, 19(1):27--39, 2018.

\bibitem{CNNSemanticDeep}
B.~Zhou, A.~Khosla, A.~Lapedriza, A.~Oliva, and A.~Torralba.
\newblock Object detectors emerge in deep scene cnns.
\newblock {\em In {ICLR}}, 2015.

\bibitem{interpretableDecomposition}
B.~Zhou, Y.~Sun, D.~Bau, and A.~Torralba.
\newblock Interpretable basis decomposition for visual explanation.
\newblock {\em In ECCV}, 2018.

\end{thebibliography}
}

\end{document}